\documentclass[journal,twoside,web]{ieeecolor}
\usepackage{jsen}
\usepackage{cite}
\usepackage{amsmath,amssymb,amsfonts}
\usepackage{algorithmic}
\usepackage{graphicx}
\usepackage{textcomp}
\usepackage{wrapfig}
\usepackage{textgreek}

\usepackage{booktabs}
\usepackage{textcomp}
\usepackage{siunitx}
\usepackage{graphicx}
\usepackage{cite}
\usepackage{multirow}
\usepackage{mathtools}
\usepackage{color}

\def\BibTeX{{\rm B\kern-.05em{\sc i\kern-.025em b}\kern-.08em
    T\kern-.1667em\lower.7ex\hbox{E}\kern-.125emX}}
\definecolor{abstractbg}{rgb}{0.89804,0.94510,0.83137}
\begin{document}

\title{IMU Based Deep Stride Length Estimation With Self-Supervised Learning}
\author{Jien-De Sui and Tian-Sheuan Chang, \IEEEmembership{Senior Member, IEEE}
\thanks{This work was supported in part by the Ministry of Science and Technology, Taiwan, under Grant 109-2634-F-009-022.}
\thanks{Jien-De Sui is with the Institute of Electronics,
National Chiao Tung University, Taiwan (e-mail: vigia117.ee06g@nctu.edu.tw). }
\thanks{Tian-Sheuan Chang is with the Institute of Electronics,
National Chiao Tung University, Taiwan (e-mail: tschang@g2.nctu.edu.tw).}
\thanks{
© 2021 IEEE.  Personal use of this material is permitted.  Permission from IEEE must be obtained for all other uses, in any current or future media, including reprinting/republishing this material for advertising or promotional purposes, creating new collective works, for resale or redistribution to servers or lists, or reuse of any copyrighted component of this work in other works.\\
J. -D. Sui and T. -S. Chang, "IMU Based Deep Stride Length Estimation With Self-Supervised Learning," in IEEE Sensors Journal, vol. 21, no. 6, pp. 7380-7387, 15 March, 2021, doi: 10.1109/JSEN.2021.3049523.
}
}

\IEEEtitleabstractindextext{%
\fcolorbox{abstractbg}{abstractbg}{%
\begin{minipage}{\textwidth}%
\begin{wrapfigure}[15]{r}{3.2in}%
\includegraphics[width=3in]{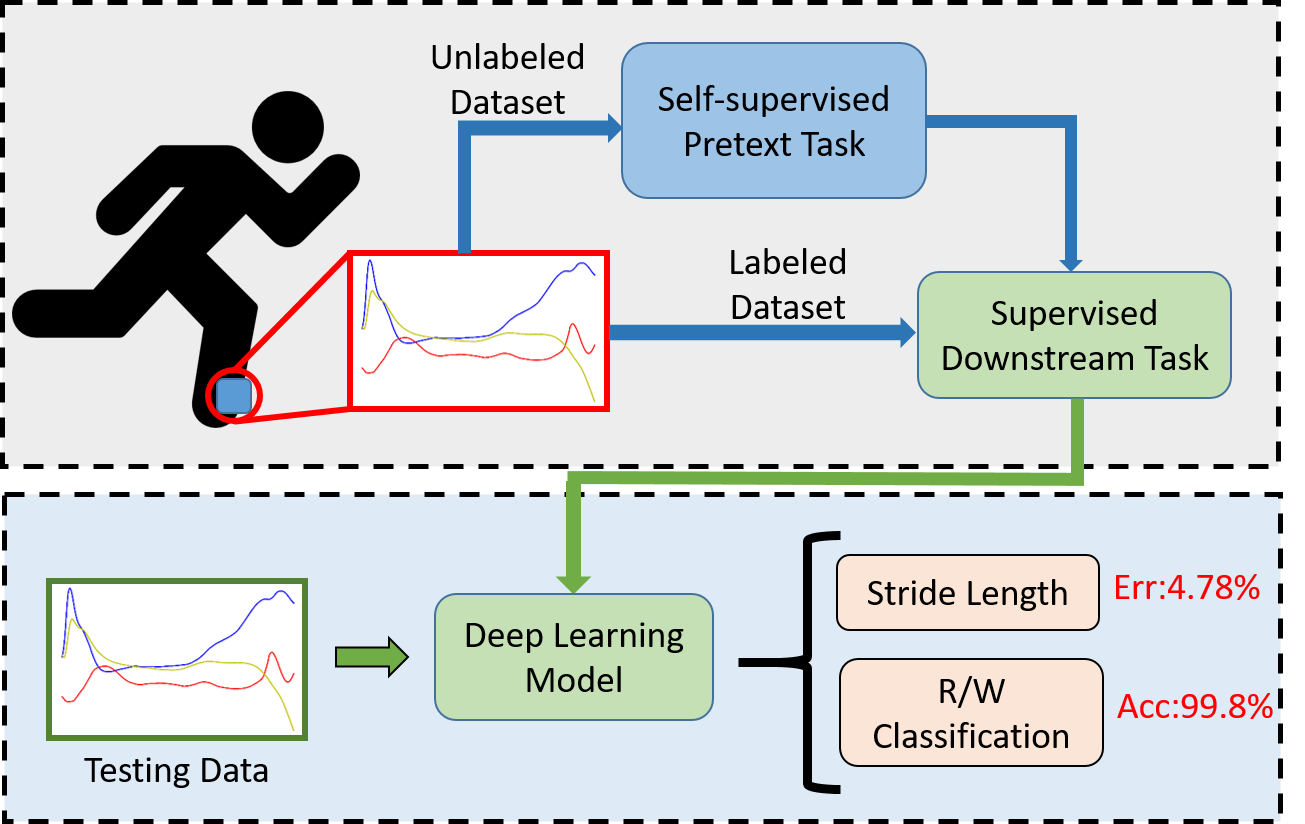}%
\end{wrapfigure}%
\begin{abstract}
Stride length estimation using inertial measurement unit (IMU) sensors is getting popular recently as one representative gait parameter for health care and sports training.  The traditional estimation method requires some explicit calibrations and design assumptions. Current deep learning methods suffer from few labeled data problem. To solve above problems, this paper proposes a single convolutional neural network (CNN) model to predict stride length of running and walking and classify the running or walking type per stride. The model trains its pretext task with self-supervised learning on a large unlabeled dataset for feature learning, and its downstream task on the stride length estimation and classification tasks with supervised learning with a small labeled dataset. The proposed model can achieve better average percent error, 4.78\%, on running and walking stride length regression and 99.83\%  accuracy on running and walking classification, when compared to the previous approach, 7.44\% on the stride length estimation.

\end{abstract}

\end{minipage}}}

\maketitle

\section{Introduction}
\label{sec:introduction}
\IEEEPARstart{S}{tride} length is an important gait parameter widely used in personal health monitoring systems \cite{jylha2001walking} \cite{woo1999walking} \cite{purser2005walking} for sports training and daily life. For daily life, the accurate walking distance and average velocity can represent the energy expenditure. For sports training, tracking running/jogging performance from distance and velocity can prevent overtraining and provide more effective training \cite{runningspeedInflu} \cite{runniningEffects}. Above tracking applications with global positioning system (GPS) have high energy consumption and need strong satellite signal strength. Even more, it only provides average speed and total distance instead of single step speed and length. In contrast, an IMU system has low power consumption and flexible usage to obtain per stride information. This paper adopts a foot-mounted IMU device instead of the smartphone built-in IMU for data collection since a dedicated IMU device is much lighter than a smartphone. A lightweight device is much desired for high speed running in the athletes training to avoid carry extra weight or equipment.

Various IMU based approaches have been proposed in recent years to extract stride length, either with conventional signal processing or deep learning methods, but these approaches still have some shortcomings. Lateddo \cite{ladetto2000foot} developed a linear combination formula with step frequency and acceleration variance. Weinberg \cite{weinberg2002using} used vertical acceleration on the upper body of the subject to model the stride length with an adjustable parameter to adjust the distance. Kim \cite{kim2004step} proposed an equation that uses average acceleration in each step during walking.  Yao \cite{yao2020robust} developed an equation consisting of three trainable step parameters, step frequency, and acceleration to estimate walking stride length with a smartphone. The above researches only focus on the walking stride length. Shin \cite{shin2007adaptive}
 indicates that the stride length of walking and running can not be estimated with the same regression model. Therefore, they used two regression equations based on step frequency and acceleration variance to estimate walking and running stride length respectively.  However, the above approaches usually need calibrations or explicit filter processing to get an accurate result, which is time-consuming and not universal for different situations. 

Beyond the signal processing approaches, another approach is to use a deep learning model to predict stride length due to their better performance than their conventional counterparts. Hannink\cite{hannink2016sensor} proposed a CNN model to predict walking stride length. Sui\cite{de2019deep} trained a fused convolutional neural network to extract gait trajectory in 3-D domain. Chen\cite{ChenAllConv} used an all convolutional neural network to fit walking stride length. Above deep learning researches only train and test on walking data, and fail to be extended to the running data case. Zhang \cite{zhang2019accurate}  presented a machine learning based method to predict walking and running stride length with not only an embedded IMU sensor but also a multi-cell piezo-resistive insole, but they did not test on the high-speed running data and only verified their model on the treadmill instead of the over ground data. Above deep learning or machine learning methods use supervised training, which usually requires massive amount of labeled data to get better performance.  However, the human-annotated labels are both expensive and time consuming, especially in this kind of sensor data collection.

To address above issues, this paper proposes a stride length estimation model that uses self-supervised learning on a large unlabeled dataset to train a pretext task and then uses supervised learning with a small labeled dataset to train a downstream task. Self-supervised training has been getting popular recently to improve performance when labeled data are scarce but with abundant unlabeled data\cite{selfsupervisedRotation} \cite{dahlkamp2006self} \cite{noroozi2016unsupervised} \cite{doersch2015unsupervised}\cite{lan2019albert}\cite{jawed2020self}. For stride length estimation, \cite{gu2018accurate} used stacked autoencoders to predict walking and jogging stride length, which used unlabeled data to learn features. But their model just used three layers of encoders and decoders, which is too simple to have a good performance for high speed running case as the reproduced result shown in Section IV. In contrast, the proposed model is able to estimate stride length for running and walking at different speeds and per stride running and walking classification. The self-supervised learning reconstructs sensor data as a self learning task with different data transformations on unlabeled data to learn representative features of the sensor data. The supervised learning augments the small label dataset by adding random noise for better model training, and adopts a percentage error weighted root mean squre error (PEW\_RMSE) as loss function to overcome the unbalance of ground truth range between walking and running. The final evaluation shows significant improvements (7.44\% to 4.78\%) compared to the previous approach.

The rest of the paper is organized as follows. Section II shows the overall sensor system used in this paper. Section III presents our network and training methodology. The evaluation results and comparisons are shown in Section IV. Finally, we concluded this paper in Section V.

\section{Sensor system}
\label{sec:methodology}

 \begin{figure}[tb]
\centering

    \includegraphics[width=80mm,scale=0.5]{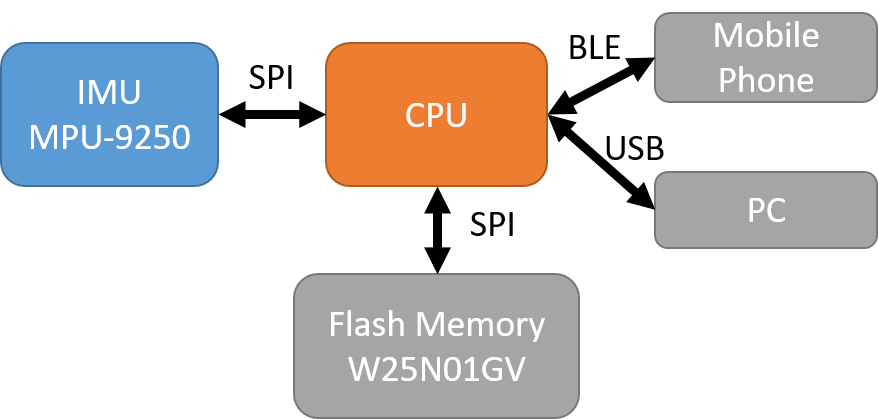}
    \caption{The sensor system architecture. The system can connect to PCs and mobile phones via USB and Bluetooth.} 
    \label{fig.sensor_flow}
\end{figure}

 \begin{figure}[tb]
\centering

    \includegraphics[width=80mm,scale=0.5]{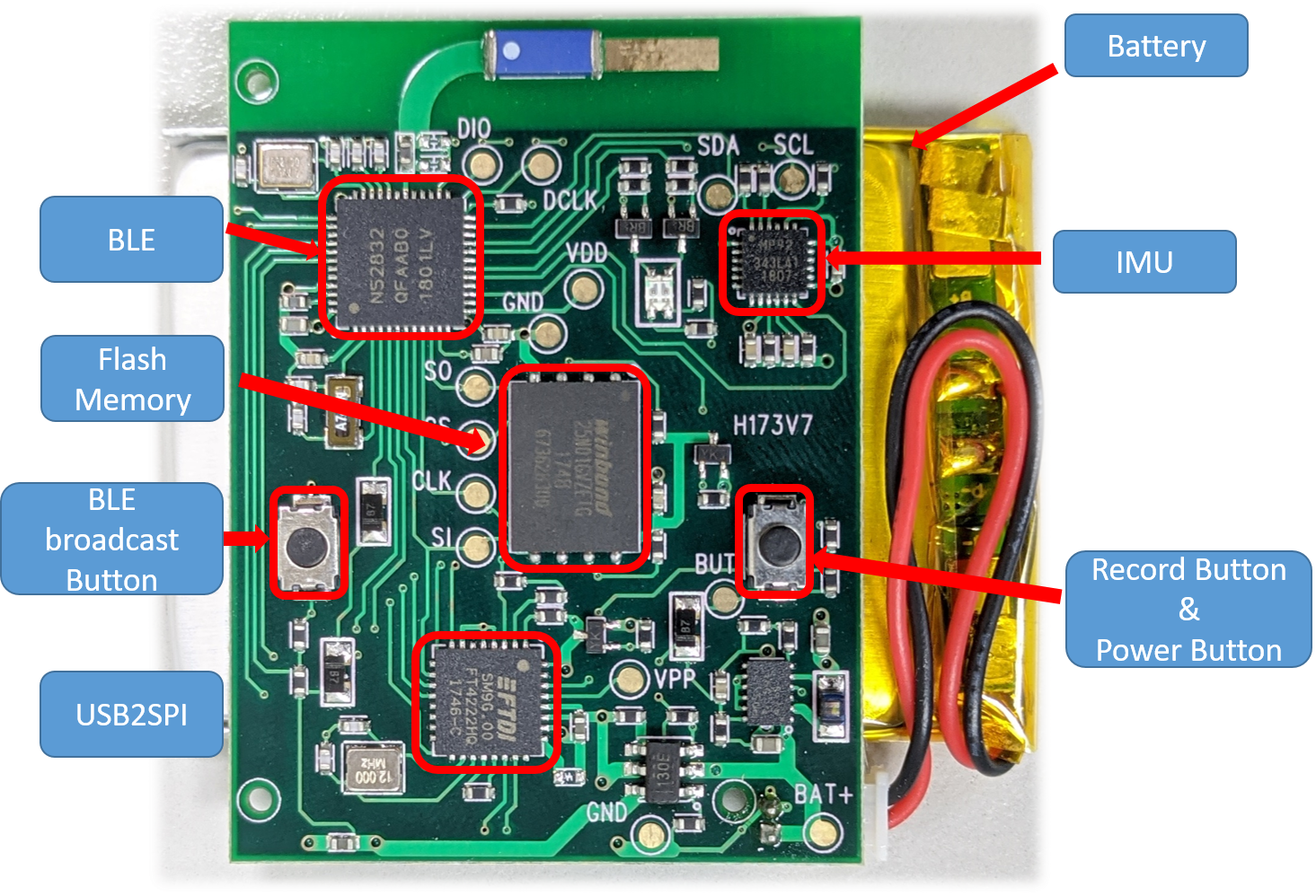}
    \caption{The sensor PCB.}
    \label{fig.sensor_arc}
\end{figure}

\subsection{Sensor}

 Fig. \ref{fig.sensor_flow} shows the sensor system architecture to collect our dataset. It consists of a consumer grade 9-axis IMU from InvenSense MPU-9250 (tri-axis accelerometer, tri-axis gyroscope, and tri-axis magnetometer), and has a 1-Gb flash memory for storage. For data transmission, this system can transmit sensor data to a computer via USB for further data processing or connect to a mobile phone via Bluetooth for real-time calculation. \par 
The whole system is integrated in a 30 mm $\times$ 30 mm printed circuit board (PCD), as shown in Fig. \ref{fig.sensor_arc}. This sensor consumes only 9.3 mA current for the highest sampling rate (1000 Hz) and can ensure all day
long continuous data recording. The sampling rate can be 1000, 500, 200,
100, and 50Hz. In this research, we used 1000Hz sampling rate to collect our dataset. The accelerometer has a dynamic range of ± 16 g
(g = 9.81 m/$s^{2}$) and the gyroscope has a dynamic range of ±
2000 °/ s\cite{NAXSENmanual}. Table \ref{tab.sensor} shows the details of this sensor.

\renewcommand\arraystretch{1.37}
\begin{table}[tb]
\caption{The detail of the sensor.}
\centering
\scalebox{0.9}{
\begin{tabular}{cccc}
\toprule

                                                        & PCB                                                                            & Processor                                                                 & BLE                                                                      \\ \hline
Spec.                                                   & 30 mm $\times$ 30 mm                                                                    & \begin{tabular}[c]{@{}c@{}}32-bit ARM \\ Cortex-M4 SoC\end{tabular}       & \begin{tabular}[c]{@{}c@{}}Nordic Semiconductor \\ nRF52832\end{tabular} \\ \toprule \toprule
                                                        & Storage                                                                        & Data rate                                                                 & Current                                                                  \\ \hline
Spec.                                                   & \begin{tabular}[c]{@{}c@{}}Windbond W25N01GV \\ 1G-bit NAND flash\end{tabular} & \begin{tabular}[c]{@{}c@{}}1000Hz (highest)\\20Hz (lowest)\end{tabular} & \begin{tabular}[c]{@{}c@{}}9.3 mA (at 1000Hz)\\ 4.2mA (at 20Hz)\end{tabular}               \\ \toprule \toprule
                                                        & \begin{tabular}[c]{@{}c@{}}tri-axis\\ accelerometer\end{tabular}               & \begin{tabular}[c]{@{}c@{}}tri-axis\\ gyroscope\end{tabular}              & \begin{tabular}[c]{@{}c@{}}tri-axis\\ magetometer\end{tabular}           \\ \hline
\begin{tabular}[c]{@{}c@{}}Dynamic\\ Range\end{tabular} & ±16 g                                                                           & ±2000°/s                                                                  & ±4800 µT                                                             \\ \bottomrule
\end{tabular}
}
\label{tab.sensor}
\end{table}
\subsection{Data preprocessing}

 \begin{table*}[]
 \caption{The detail of the labeled dataset and unlabeled dataset.}
\centering
\begin{tabular}{ccccccccc}
\toprule
\multicolumn{9}{c}{Labeled Dataset}                                                                                                            \\ \hline
\multicolumn{9}{c}{3 Subjects}                                                                                                               \\ \hline
Type   & Walk 5km/hr   & Walk 7km/hr   & Run 9km/hr    & Run 11km/hr    & Run 13km/hr     & Run 15km/hr    & Run 17km/hr    & Run 19km/hr    \\
Steps   & 180           & 360           & 360           & 360            & 360             & 360            & 240            & 120            \\
Seconds & 17.62          & 30.77          & 22.98          & 26.88           & 25.56            & 24.80           & 14.84           & 8.45           \\ \bottomrule \toprule
\multicolumn{9}{c}{Unlabeled Dataset}                                                                                                          \\ \hline
\multicolumn{9}{c}{11 Subjects}                                                                                                              \\ \hline
Type   & \multicolumn{2}{c}{treadmill} & \multicolumn{2}{c}{playground} & \multicolumn{4}{c}{asphalt road   ( flat \& incline \& decline )} \\ 
Steps   & \multicolumn{2}{c}{3220}      & \multicolumn{2}{c}{12414}    & \multicolumn{4}{c}{30637}                                          \\
Seconds & \multicolumn{2}{c}{2664}      & \multicolumn{2}{c}{9198}      & \multicolumn{4}{c}{25560}                                            \\ \bottomrule
\end{tabular}
\label{tab.dataset}
\end{table*}

\subsubsection{Data collection}

\begin{figure}[tb]
    \includegraphics[width=\linewidth]{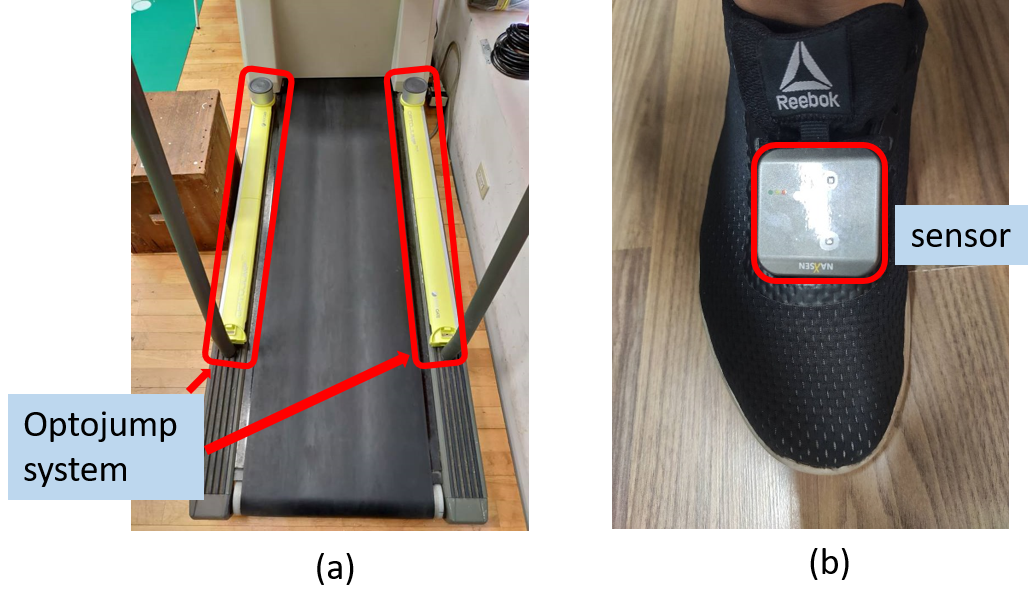}
    \caption{(a) The position of the Optojump system, which is placed on both sides of the treadmill. (b) The position of the sensor system, which is is mounted and fixed on the shoes.} 
    \label{fig.sensor_position}
\end{figure}

For the labeled dataset, we collected data of normal subjects, including two males and one female, aged between 20 and 30 years old. To collect the labeled data, we used an optical measurement - Optojump system, which contains one pair of the receiving and transmitting bars that can record the position and time of the contact point. Therefore, we can extract stride length by this system. But due to the complicated setting, calibration, and length of the bars, this system is only suitable for indoor short-distance use. To overcome the distance problem, we use a treadmill to help collect long-distance sensor data.  The experiments include running and walking at various speeds (5, 7, 9, 11, 13, 17 and 19 km/h) based on the treadmill record and forefoot and rearfoot strike patterns. Fig. \ref{fig.sensor_position} shows the placements of the Optojump and sensor systems.

\begin{figure}[t]
\centering
    \includegraphics[width=90mm,scale=0.5]{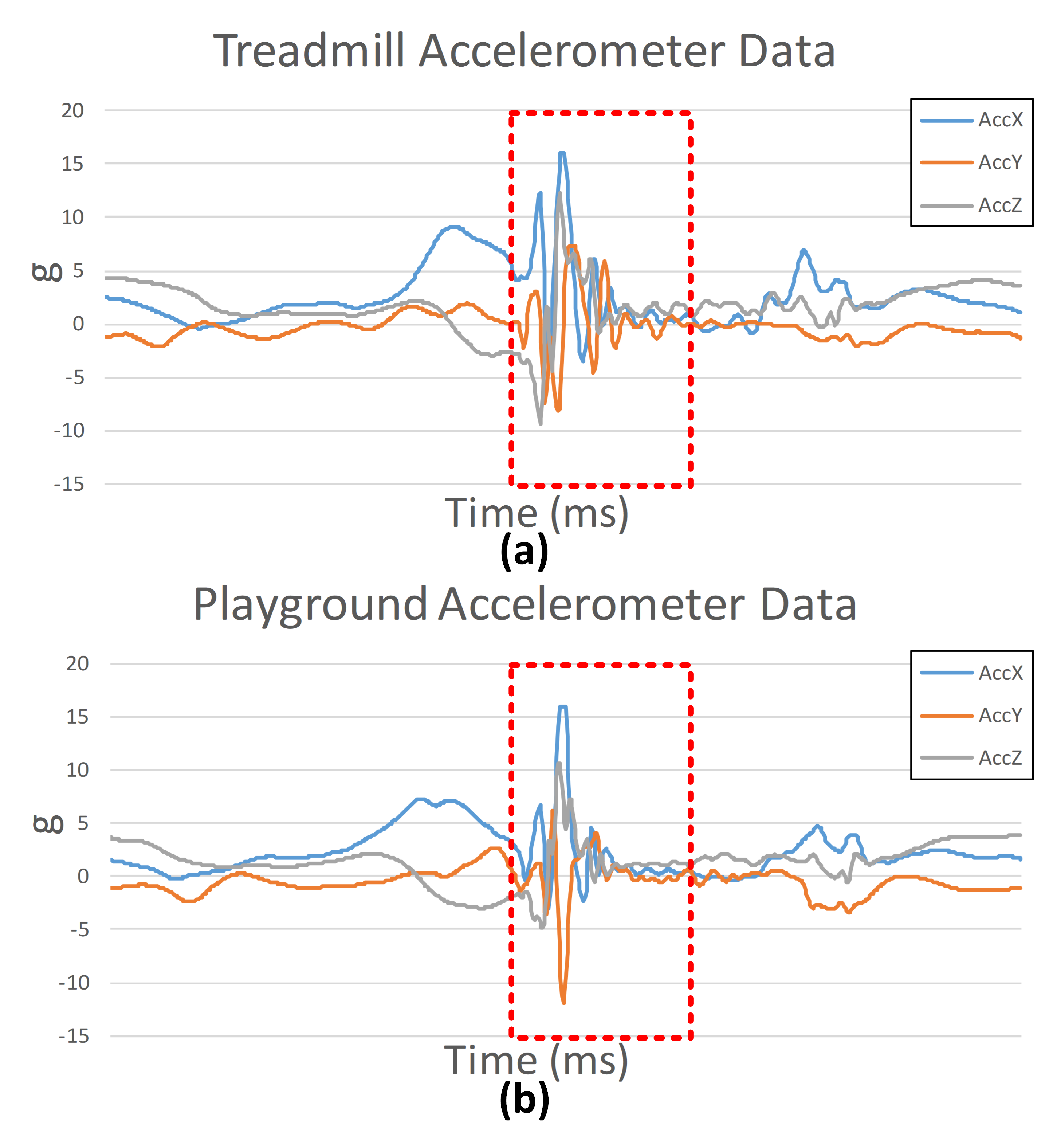}
    \caption{The accelerometer data of the same subject with different ground materials. }
    \label{fig.different_ground}
\end{figure}

 For outdoor unlabeled data, we collect a more diverse unlabeled dataset, including different persons, action patterns, grounding, etc. The experiments collected data from 11 normal subjects. The action pattern included sprinting, jogging, normal running, and walking patterns. In addition to the action pattern, the unlabeled data contained more diverse ground materials and states, including treadmill, playground, asphalt road, incline road, and decline road. Fig. \ref{fig.different_ground} shows accelerometer data of the same subject with different ground materials. Data from the treadmill has a stronger bouncing effect and thus larger acceleration data due the more flexible belt material and structure of the treadmill, when compared to the more rigid case in the playground and asphalt road. The length of this unlabeled dataset is 624 minutes, which is 220 times longer than the labeled dataset. We use this unlabeled data with some augmentations as training data for self-supervised training. Table \ref{tab.dataset} shows the details of labeled dataset and unlabeled dataset .

\subsubsection{Stride segmentation}
\label{sec:stride_seg}
For stride length prediction, the first thing to do is stride segmentation. The ground truth of the segmentation is provided by the Optojump system, including the stride length, and start points of the stance and swing phases. Each stride will be zero padded to 600-sample-point to cover the length of one stride.\par

The collected sensor dataset is also split into one-stride long data by our algorithm. To segment sensor data, many traditional methods use threshold based methods to segment strides, but these threshold based methods are not suitable for running due to larger variations. Besides, they need extra filter processing and explicit calibration for reliable prediction. Therefore, we use our previous proposed deep learning based segmentation model, IMU-Net \cite{IMUNET2}, for stride segmentation. IMU-Net can classify each data point and is suitable for running and walking at different speeds. The accuracy of IMU-Net can achieve 99.97\% of stride detection.

\subsubsection{Data order and normalization}
Following segmentation, we can arrange IMU data into two types of order for deep learning model, namely \emph{Spatial-First} and \emph{Temporal-First}. The \emph{Temporal-First} order treats  different axis of sensor data as channels, 6 $\times$ 1 $\times$ 600 (sensors $\times$ channel $\times$ one segment length). The \emph{Spatial-First} order puts different axis of sensor data together in one channel, 1 $\times$ 6 $\times$ 600 (channel $\times$ sensors $\times$ one segment length). In this paper, we use the \emph{Spatial-First} order since \emph{Spatial-First} order is better than \emph{Temporal-First} order on the IMU data \cite{IMUNET2} .

 \begin{figure*}[tb]
    \includegraphics[width=\linewidth,keepaspectratio]{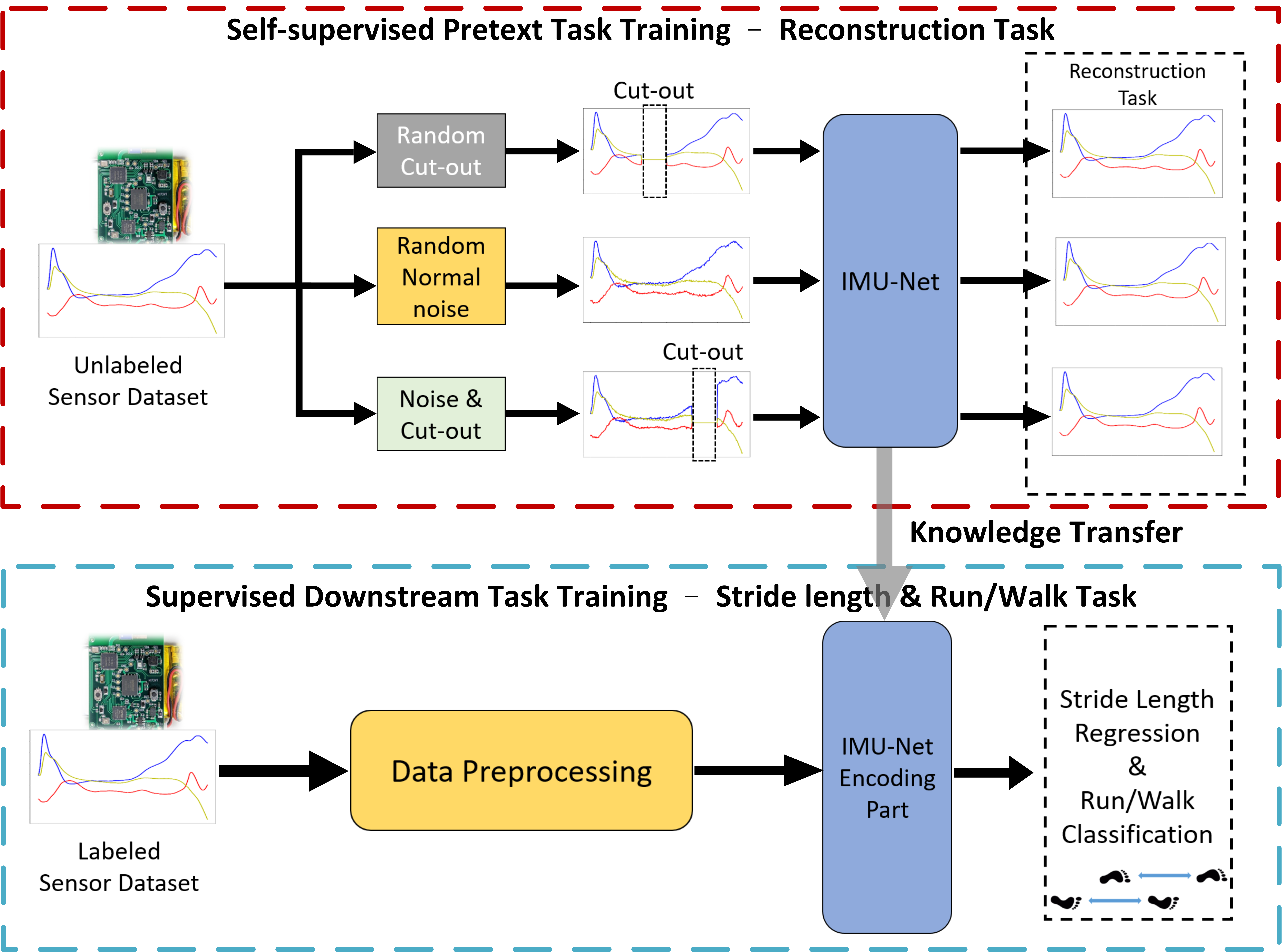}
    \caption{Illustration of the whole process. The entire process is divided into self-supervised reconstruction pretext task and supervised downstream tasks, run/walk classification and stride length estimation. The pretext task uses three data augmentation methods on the training data as input to train the IMU-Net with data reconstruction for self-supervised learning. Then, only the encoding part of IMU-Net is extracted with a downstream network to solve the target tasks.  }
    \label{fig.selfsupervised}
\end{figure*}

Finally, we apply normalization to our dataset. In our dataset, the range of accelerometer and gyroscope is different, ± 16 g (g = 9.81 m/$s^{2}$) and ± 2000 °/s respectively. To avoid accelerometer or gyroscope data dominating in the whole process, we divide these data by their maximum value so that the data is within ± 1 in every axis of sensor data.

\section{Deep learning training }
Fig. \ref{fig.selfsupervised} shows the whole training processes that can be divided into two parts. One is self-supervised feature learning, which reconstructs sensor data as a pretext task. The other part is supervised learning for downstream tasks, which uses a labeled dataset for stride length estimation and run/walk classification.

\subsection{Self-supervised reconstruction pretext task}
\subsubsection{Reconstruction network}\label{sec:Reconstruction_network}

The self-supervised network uses IMU-Net \cite{IMUNET2}, which is modified from U-Net \cite{Unet} with the same structure but fewer channels. U-Net contains an encoding path and a decoding path. The encoding path consists of several combinations of double-convolutional layer and max-pooling layers. The decoding path is the reverse of the encoding path, where each layer concatenates upsampled feature maps with the corresponding feature maps from the encoding path. In the last layer, both IMU-Net and U-Net use a convolutional layer with softmax for classification. For the reconstruction task, we use a convolutional layer with a linear layer to map feature maps into ground truth (raw sensor data).
Compared to U-Net, IMU-Net reduces the number of filters for lower complexity and changes max-pooling size from 2 $\times$ 2 to 1 $\times$ 4 to set the receptive field covering whole IMU data input. Besides, this 1 $\times$ 4 max-pooling is to downsample data along the temporal domain rather than the spatial domain (different axis of the sensor), because the spatial domain downsampling will lose too much information.

\subsubsection{Sensor data augmentation}
Self-supervised learning uses different data augmentations to build self generated targets to learn from a large amount of non-human-labeled data. \cite{shorten2019survey} surveyed and discussed many image data augmentation, such as jittering, scaling, flipping, cropping, rotation, and color space transformations, etc..\cite{sensor_data_aug} also proposed many sensor data augmentation methods, like jittering, permutating, and time-warping methods. However, most of them are not suitable for the regression task of the time series data. This paper chooses two methods suitable for our sensor data reconstruction task, cut-out and random normal noise.\par

First, we randomly split the unlabeled data into 600-sample-point segments to increase dataset size. After the random segmentation, the data length becomes 1855 minutes, 3 times than the original. We will use these augmented sensor data as self generated target. \par

\begin{figure}[tb]
    \includegraphics[width=\linewidth]{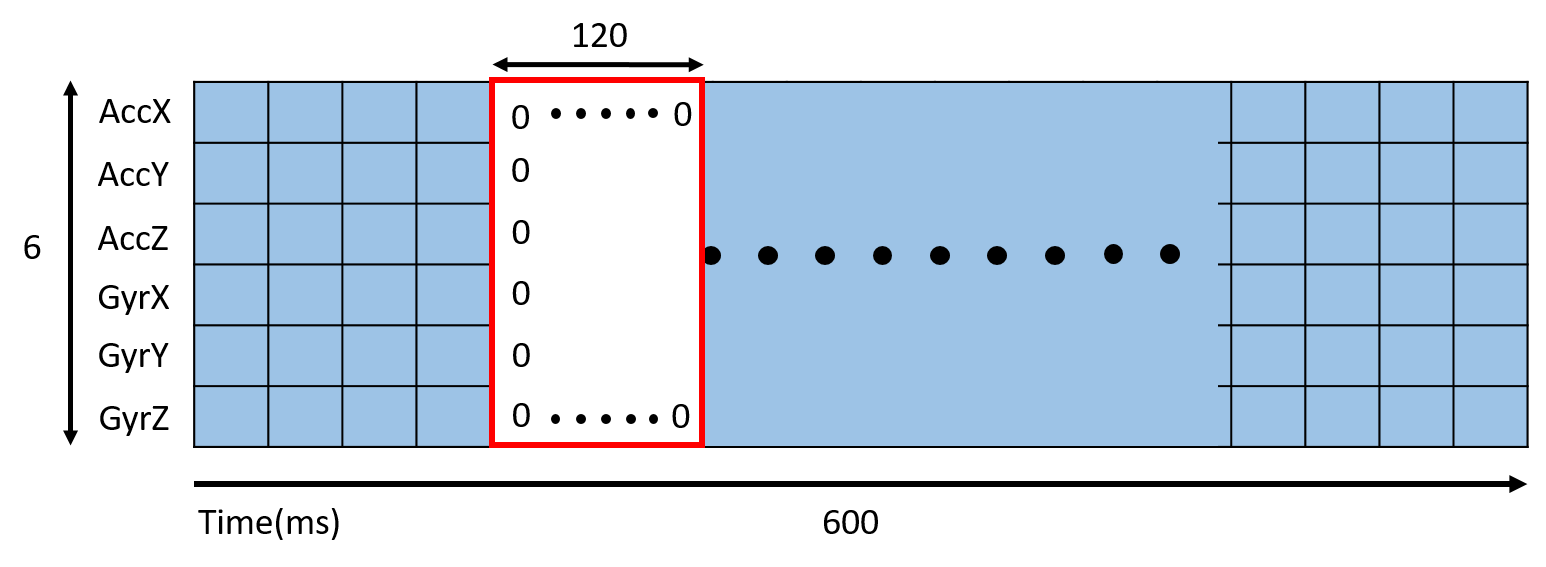}
    \caption{The cut-out data augmentation. We cut out the one-fifth of the sensor data (120-sample-point) and fill it with zero for data augmentation.}
    \label{fig.cutout}
\end{figure}

The cut-out augmentation randomly removes part of the sensor data and fills in that part with zero for data reconstruction task. The cut-out size is one-fifth of the sensor data (120-sample-point), as shown in Fig. \ref{fig.cutout}. 

\begin{figure}[tb]
    \includegraphics[width=\linewidth]{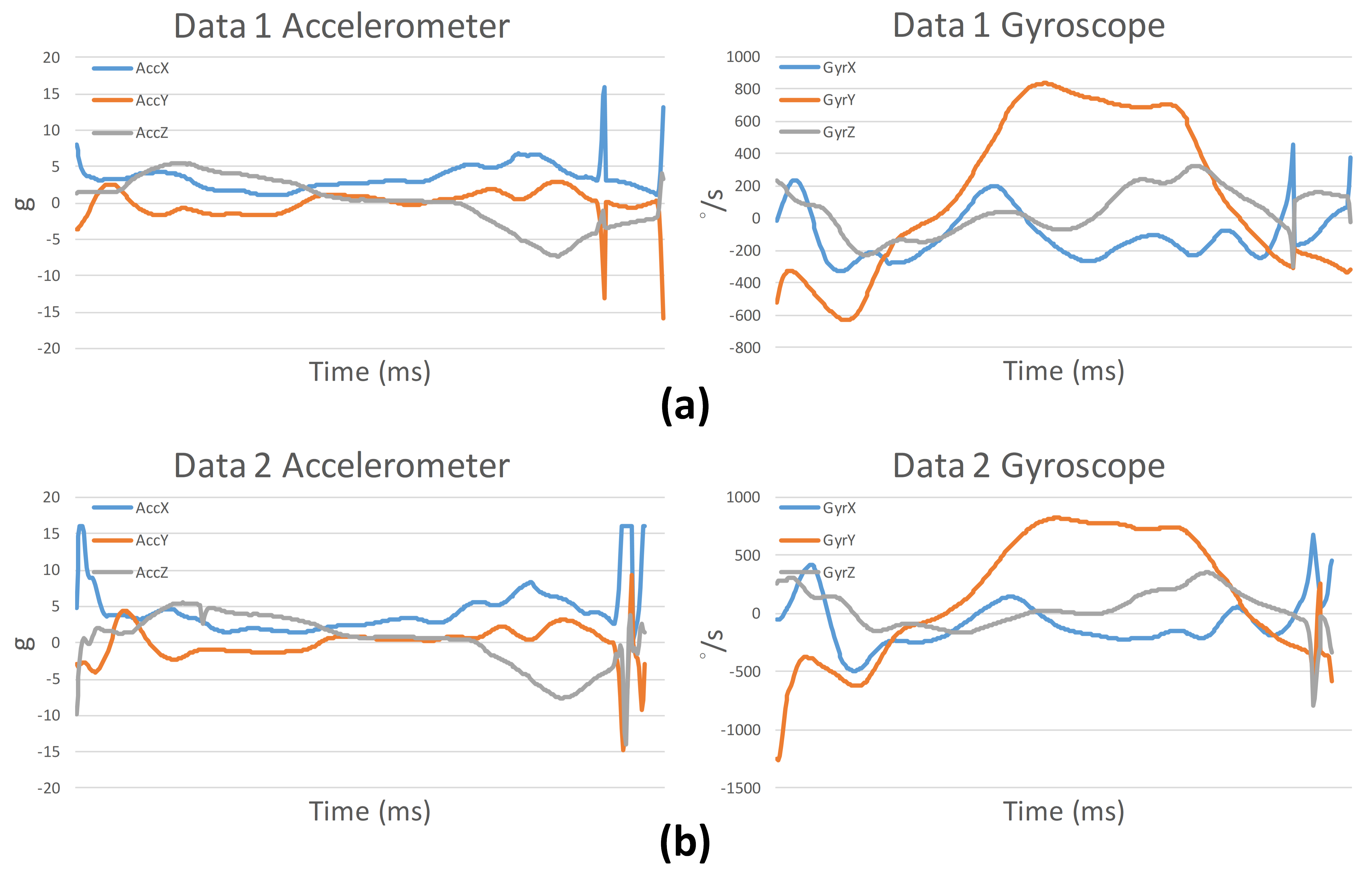}
    \caption{Different sensor data with same stride length ground truth data, 231 cm.} 
    \label{fig.sameground}
\end{figure}

The random normal noise augmentation adds zero mean random normal noise on both accelerometer data and gyroscope data. The random noise is to simulate the effect of the sensor noise. In our dataset, we found that many sensor data have different waveforms but with the same stride length, as shown in Fig. \ref{fig.sameground}. The example waveforms in Fig. \ref{fig.sameground} is very similar but will get very different prediction outcome. Thus, slight noise will cause a great error on prediction. This result means that the model is too sensitive to noise. \cite{wang2019pedestrian} presented a denoising autoencoder approach to reconstruct their merged feature and solve the noise problem. In this paper, instead of the denoising approach, we add random noise to the training dataset, forcing the network to extract more significant features. This makes the model more noise tolerant and generates more stable prediction results. The standard deviation of the noise is set as one percent of the maximum value of the sensor data, that is, 0.16 for accelerometer data and 20 for gyroscope data, as shown in \eqref{eq:noise}. We also combine these two augmentation methods to create more diverse training data, as shown in Fig. \ref{fig.selfsupervised}. Note that, we also use random normal noise augmentation on labeled sensor dataset to increase diversity of the labeled dataset.

\begin{equation}
\hat{y}_{i}^{ax}= \begin{cases} 
y_{i}^{ax} + N(\mu = 0, \sigma = 0.16), 
              &\text{if}\ ax = Acc \\ 
            y_{i}^{ax} + N(\mu = 0, \sigma = 20), 
            &\text{if}\ ax = Gyr \end{cases}\label{eq:noise}
\end{equation}

\subsubsection{Training of self-supervised learning}
Self-supervised learning uses IMU-Net to learn the features of the time series data with reconstruction as the learning task. This model uses Mean-Square-Error of the reconstructed data and self generated target as the loss. The model is trained for 100 epochs with batch size 30 with the Adam optimizer \cite{Adam}. After training, this model learns the characteristics of sensor data and can be effectively applied to our sensor tasks.

\subsection{Supervised stride length task}

\subsubsection{Network architecture}
\begin{figure}[tb]
 \centering
    \includegraphics[width=70mm,scale=0.5]{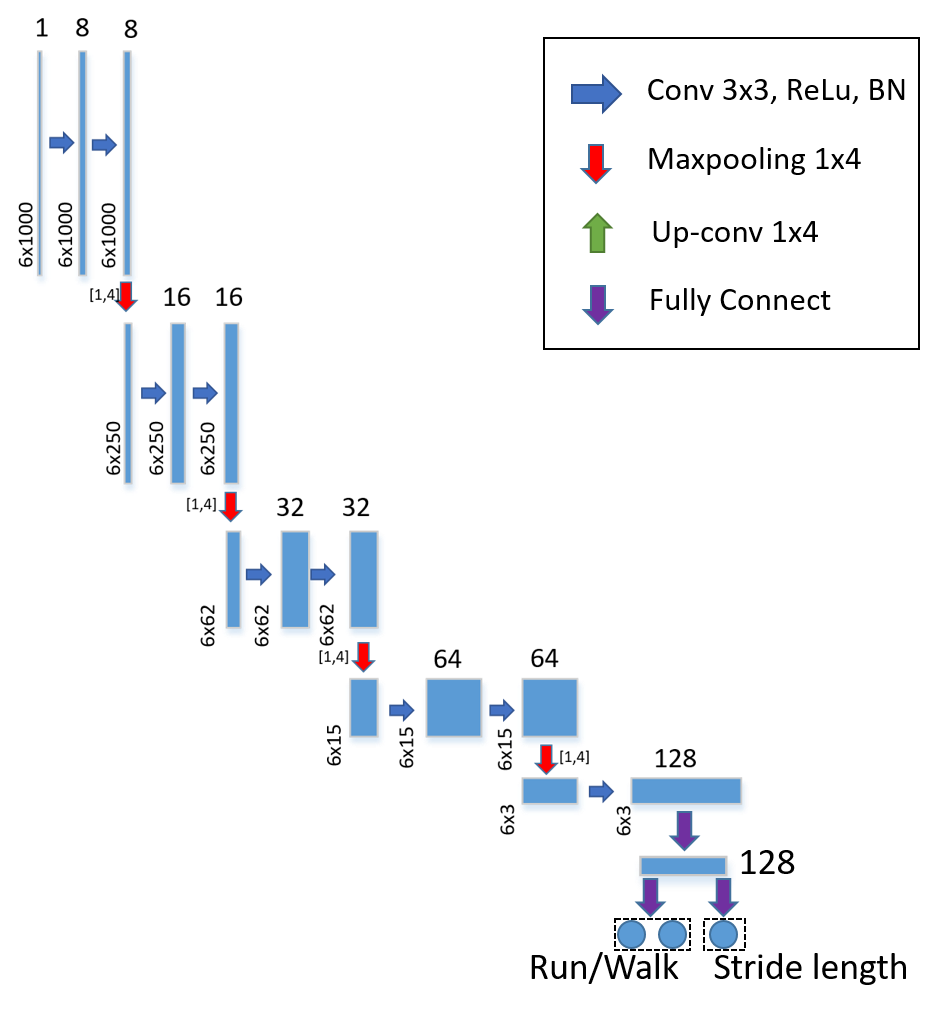}
    \caption{Supervised network architecture that uses the encoding part of IMU-Net and connects to one fully connected layer and two output layers for multi-task prediction.  } %caption是图片的标题
    \label{fig.endcoding}
\end{figure}

The downstream supervised task uses the encoding path of the trained IMU-Net as the pretrained model, and adds one fully connected hidden layer with 128 neurons and two output layers, one for run/walk classification with softmax activation function and one for the stride length regression, as shown in Fig. \ref{fig.endcoding}. This multi-task model can help solve our stride length task since the stride length of running is larger than the length of walking.

\subsubsection{PEW\_RMSE: Percent Error Weighted RMSE loss function}

For stride length estimation, we use RMSE as our initial regression loss function. However, stride length range is quite different for running and walking, from 80 cm to 140 cm for walking, and 140 cm to 300 cm for running. Thus, we will not only focus on the absolute error but also on percentage error. The percentage error is scale independent to help running/walking loss balance. However, this loss function will makes the prediction of walking stride length more accurate than the running stride length in terms of percentage error. The percentage error of the running case is larger than that in the walking by 2\%. Thus, we modify the RMSE loss function by adding the percentage error as a loss weighting as below. \par

\begin{equation}
PEW\_RMSE =  \sqrt{\frac{1}{M} \sum_{n=1}^{M}  \beta \times (p-g)^2 }   \label{eq:regression_loss}
\end{equation}

\begin{equation}
\beta = e^{\frac{\lvert p-g \rvert}{g}} \label{eq:exp_regression_loss}
\end{equation}

In the above equation, \eqref{eq:exp_regression_loss} is an exponential percentage error, which is obtained by simply dividing the prediction error by the ground truth and takes its exponential value. \eqref{eq:regression_loss} takes the percentage error into consideration, and thus it can optimize more for running that has a larger percentage error to narrow the percentage error gap between walking and running.

\subsubsection{Training and evaluation}

For the supervised multi-tasks, we train the model for 150 epochs with batch size 30, use Super-Convergence \cite{superconvergence} to speed up the training process with the Adam optimizer, and adopt cross-entropy as classification loss function and PEW\_RMSE as the regression loss function. 
For validation purposes, we apply a k-fold cross-validation approach on the training dataset and testing dataset. For this experiment, k is set to 3 for independent subjects.  We also use 10\% of the training dataset for validation. Besides, for obtaining the best model, we monitor the loss of validation set and save the best model when validation loss is minimum. The performance evaluation for the model is done by averaging the error and accuracy of all k testing dataset.

\section{Result and Analysis}

\label{Result}

\begin{table}[tb]
\caption{Average results for self-supervised only training, where the number is mean ± standard deviation.}
\begin{tabular}{cccc}
\toprule
\multicolumn{2}{c}{Metrics}                                                                     & w/o   Self-supervised & w/   Self-supervised \\ \hline
\multirow{3}{*}{MAE   (cm)}                                                           & Runing  & 13.48 ± 1.48          & \textbf{12.15 ± 3.02}         \\
                                                                                      & Walking & 10.18 ± 4.64          & \textbf{6.43 ± 3.59}          \\
                                                                                      & Avg     & 13.21 ± 0.73          & \textbf{11.41 ± 2.58}         \\ \hline
\multirow{3}{*}{\begin{tabular}[c]{@{}c@{}}Percentage\\      Error (\%)\end{tabular}} & Runing  & 6.28 ± 0.74           & \textbf{5.38 ± 1.26}          \\
                                                                                      & Walking & 7.27 ± 3.73           & \textbf{4.40 ± 2.20}          \\
                                                                                      & Avg     & 6.50 ± 0.40           & \textbf{5.24 ± 0.99}          \\ \hline
\begin{tabular}[c]{@{}c@{}}Run/Walk\\      Classification \\ (\%)\end{tabular}        & Avg     & \textbf{99.81 ± 0.32}          & 99.43 ± 0.60         \\ \bottomrule
\end{tabular}
\label{tab.self_result}
\end{table}

\subsection{Ablation study}
\subsubsection{Self-supervised only training}
\begin{figure}[tb]
    \includegraphics[width=\linewidth]{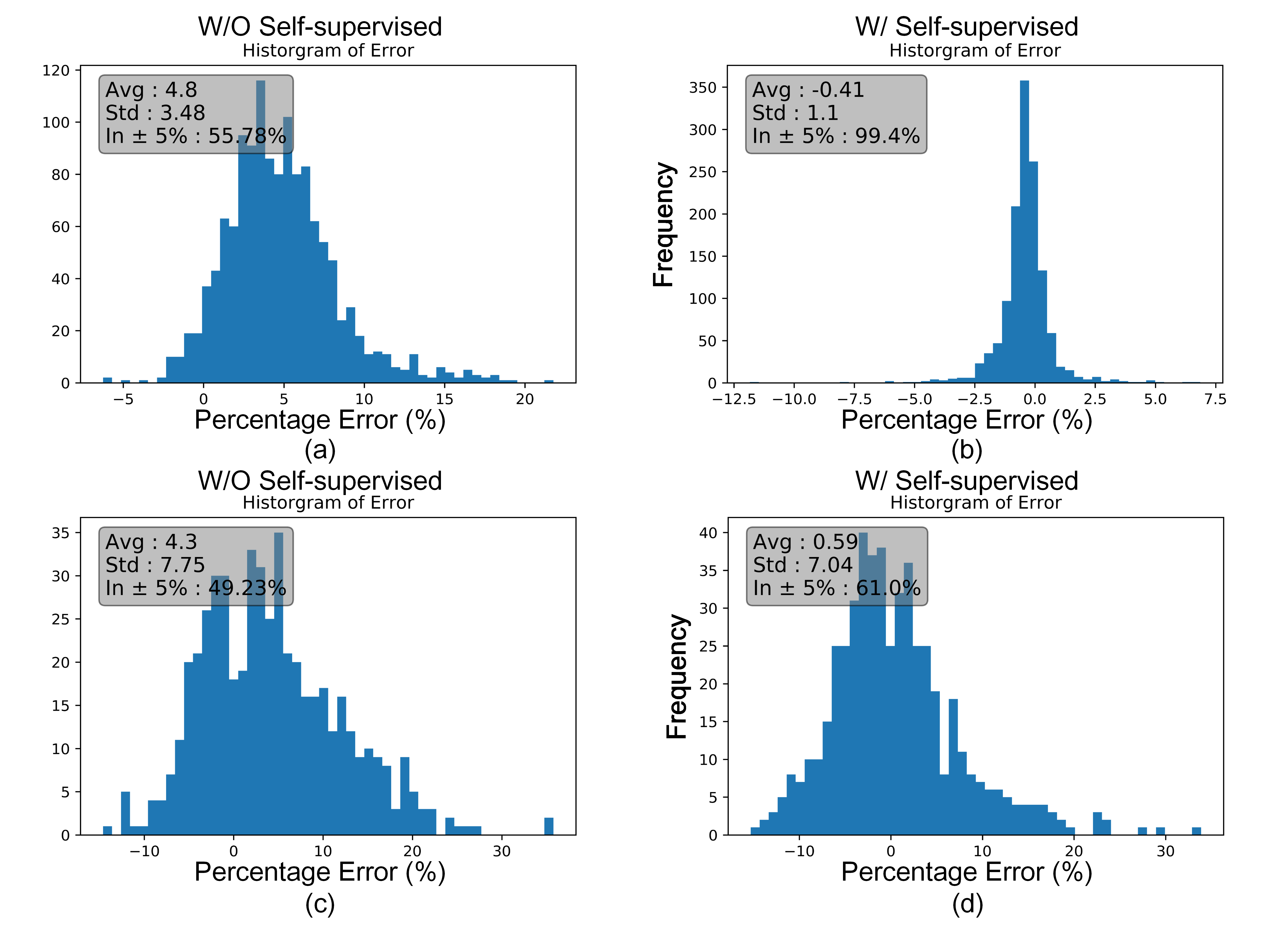}
    \caption{Histogram of error distribution.  (a) and (b) are training errors. (c) and (d) are testing errors. (a) and (c) do not use self-supervised learning and (b) and (d) use self-supervised learning. } 
    \label{fig.self_his}
\end{figure}

Fig. \ref{fig.self_his} depicted the comparison with self-supervised training or not from one fold of the evaluation process, where the “w/o self supervised” model uses the labeled data for training, and the “w self supervised” model uses the unlabeled data for pretext task training and labeled data for downstream task training. As shown in Fig. \ref{fig.self_his} (a) and (b), our self-supervised training can improve the training performance where the histogram within 5\% error is increased from 55.78\% to 99.4\%. Table \ref{tab.self_result} shows  the  mean  error  and  standard  deviation that the self-supervised training can effectively increase the model accuracy for both running and walking cases, reducing MAE error from 13.21 cm to 11.41 cm, and the percentage error from 6.5\% to 5.24\%.

\subsubsection{Loss function}
\begin{figure}[tb]
\centering
    \includegraphics[width=80mm]{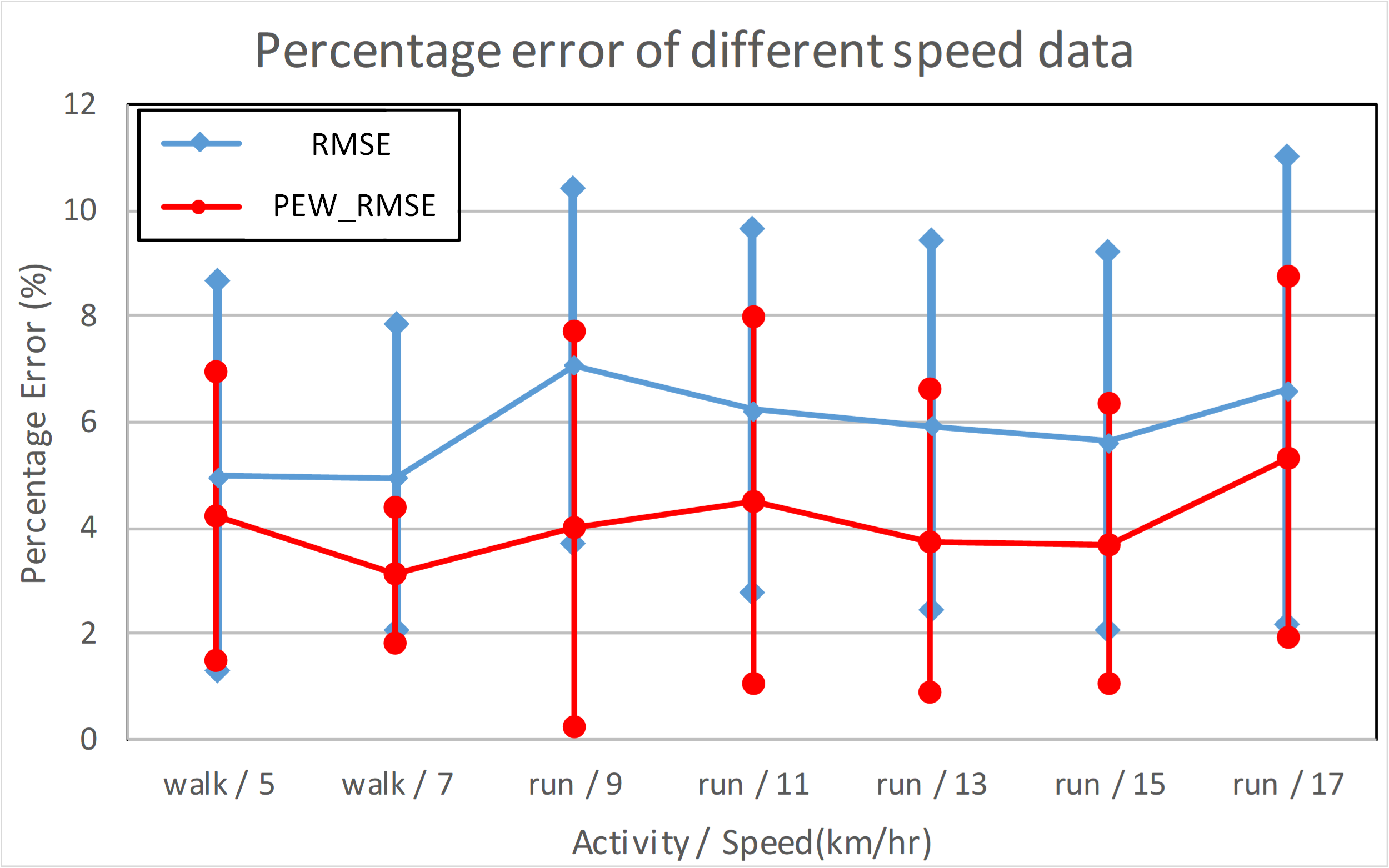}
    \caption{The percentage error for one test case.} %caption是图片的标题
    \label{fig.exp}
\end{figure}

 Fig. \ref{fig.exp} shows the effect of different regression loss functions for one test case. Other cases have similar trends. For RMSE loss, the running stride length has higher errors than the walking stride length due to much higher variation in the large and high speed movement.  With our proposed PEW\_RMSE, error in each case is reduced and also close to each other. Table \ref{tab.lossfunction_result} shows the detailed average error number, which have reduced total errors to 4.78\%.

\begin{table}[]
\centering

\caption{Comparisons of different loss functions.}
\scalebox{1}{
\begin{tabular}{cccc}
\toprule
\multicolumn{2}{l}{Metrics} & \multicolumn{1}{c}{\begin{tabular}[c]{@{}c@{}}CrossEntropy \& \\ RMSE\end{tabular}} & \multicolumn{1}{c}{\begin{tabular}[c]{@{}c@{}}CrossEntropy \& \\ PEW\_RMSE\end{tabular}} \\ \hline
\multirow{3}{*}{MAE (cm)}                                                        & Runing  & 12.15                & \textbf{10.62}     \\
                                                                            & Walking & 6.43                & \textbf{6.22}      \\
                                                                            & Avg     & 11.41                & \textbf{10.15}     \\ \hline
\multirow{3}{*}{\begin{tabular}[c]{@{}c@{}}Percentage \\ Error(\%)\end{tabular}} & Runing  & 5.38              & \textbf{4.82}    \\
                                                                            & Walking & 4.40               & \textbf{4.37}    \\
                                                                            & Avg     & 5.24               & \textbf{4.78}    \\ \hline
\begin{tabular}[c]{@{}c@{}}Run/Walk\\ Classification\end{tabular}           & Avg     & 99.43              & \textbf{99.83\%}   \\ \bottomrule
\end{tabular}
}
\label{tab.lossfunction_result}
\end{table}

\subsection{Comparison with previous work}
\begin{figure}[tb]
 \centering

    \includegraphics[width=90mm]{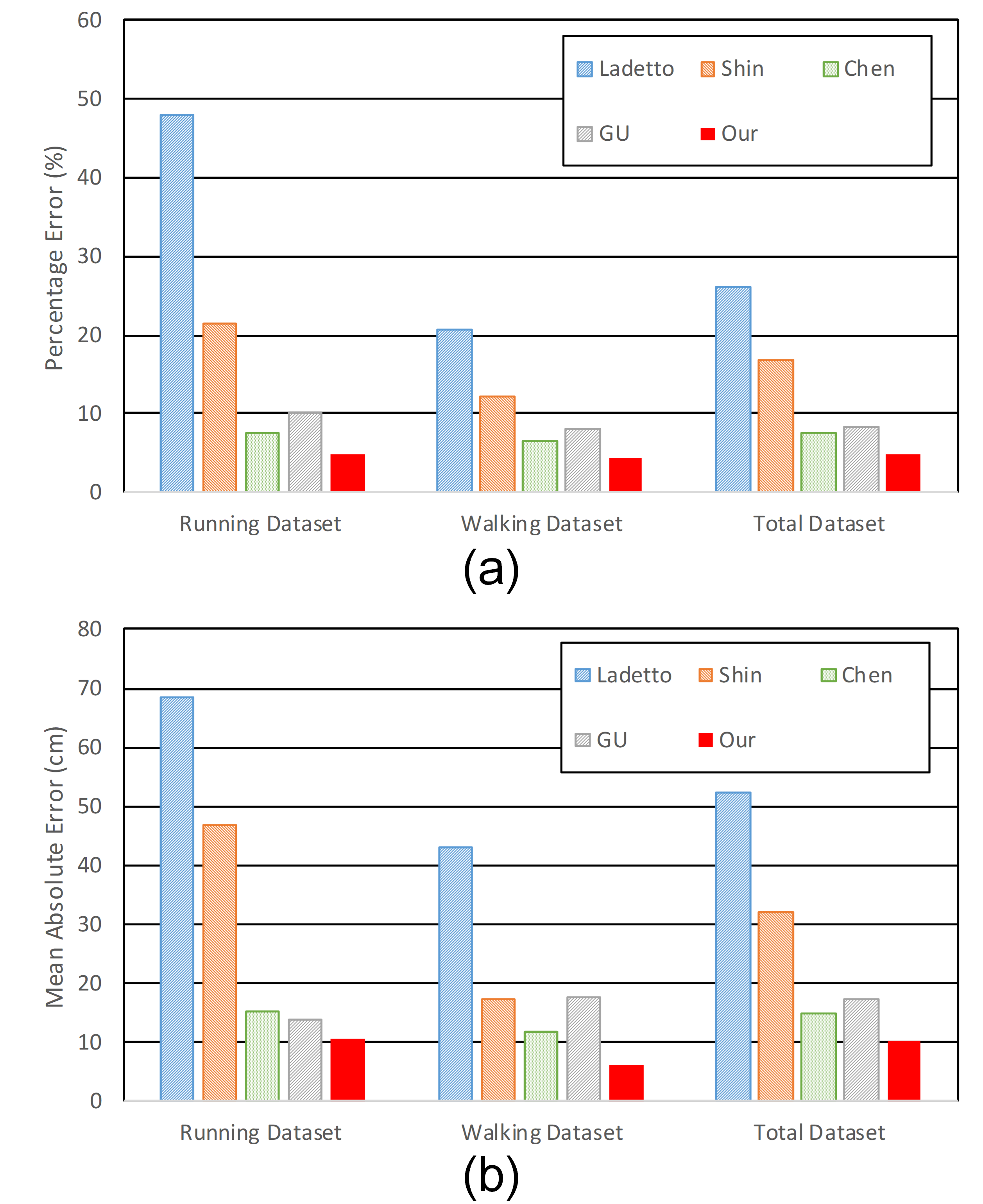}
    \caption{Comparisons with other approaches. } 
    \label{fig.compare_fig}
\end{figure}

Fig. \ref{fig.compare_fig} shows comparisons with other approaches by implementing their methods to our dataset for fair comparisons with a 3-fold cross-validation if needed. Note that, \cite{ladetto2000foot}, \cite{ChenAllConv} and \cite{gu2018accurate} only focused on walking or walking/jogging instead of high speed running data. For fair comparison, we also implement the method presented in \cite{gu2018accurate} for our data. Besides, as shown in the figure, most of the previous studies has far better prediction on walking instead of running. They are not suitable for predicting both running and walking stride length with a single model. 
A more detailed result is shown in Table \ref{tab.compare_result}. Compared to other approaches, the proposed method outperform in both running and walking conditions and achieves the lowest error rate.

\begin{table}[tb]
\caption{Comparisons with other approaches.}
\centering.
\scalebox{1}{
\begin{tabular}{cccc}
\toprule
\multicolumn{4}{c}{Percentage Error (\%)}                         \\ \hline
             & Running Dataset & Walking Dataset & Total Dataset  \\ \hline
Ladetto \cite{ladetto2000foot}      & 47.94           & 20.61           & 25.95          \\ \hline
Shin \cite{shin2007adaptive}         & 21.44           & 12.11           & 16.77         \\ \hline
Chen \cite{ChenAllConv}         & 7.56            & 6.54            & 7.44           \\ \hline
GU \cite{gu2018accurate}          & 10.06           & 8.11            & 8.37           \\ \hline
\textbf{Proposed} & \textbf{4.82}   & \textbf{4.37}   & \textbf{4.78}  \\ \bottomrule
\multicolumn{4}{c}{Mean Absolute Error (cm)}                      \\ \hline
             & Running Dataset & Walking Dataset & Total Dataset  \\ \hline
Ladetto \cite{ladetto2000foot}      & 68.46           & 43.1            & 52.2           \\ \hline
Shin \cite{shin2007adaptive}         & 46.73           & 17.38           & 32.05         \\ \hline
Chen \cite{ChenAllConv}         & 15.2            & 11.77           & 14.78          \\ \hline
GU \cite{gu2018accurate}           & 17.65            & 14.02            & 17.31          \\ \hline
\textbf{Proposed} & \textbf{10.62}  & \textbf{6.22}   & \textbf{10.15} \\ \bottomrule
\end{tabular}
}

\label{tab.compare_result}
\end{table}

\begin{table}[tb]
\caption{Model testing result on unseen data.}
\centering
\scalebox{1}{
\begin{tabular}{ccccc}
\toprule
Subject            & \begin{tabular}[c]{@{}c@{}}Ground truth\\ Distance (m)\end{tabular} & \begin{tabular}[c]{@{}c@{}}Predicted\\ Distance (m)\end{tabular} & Error (m) & \begin{tabular}[c]{@{}c@{}}Percentage\\ Error(\%)\end{tabular} \\ \hline
\multirow{2}{*}{1} & 400                                                           & 395.37                                                     & 4.63  & 1.16                                                           \\
                   & 400                                                           & 401.44                                                     & 1.44  & 0.36                                                           \\ \hline
\multirow{2}{*}{2} & 400                                                           & 375.38                                                     & 24.62 & 6.16                                                           \\
                   & 400                                                           & 378.74                                                     & 21.26 & 5.32                                                           \\ \hline
\multirow{2}{*}{3} & 400                                                           & 368.23                                                     & 31.77 & 7.94                                                           \\
                   & 400                                                           & 391.43                                                     & 8.57  & 2.14                                                           \\ \hline
Average            & 400                                                           & 385.09                                                     & 15.38 & 3.84                                                           \\ \bottomrule
\end{tabular}
}
\label{tab.unseen_data}
\end{table}

\subsection{Unseen data}
In order to verify the robustness of our model, we collected unlabeled data of three persons running or walking on the playground and only recorded the total distance, 400 meters.
First, we apply our stride segmentation on these unseen data, feed all sensor data into our trained network model, and then sum all the stride length to get the total distance to verify the accuracy. The result is shown in Table \ref{tab.unseen_data}. The proposed model can achieve 3.84\% average errors even on unseen data, which is attributed to our self-supervised learning.

\section{Conclusion}

This paper presents an IMU based stride length estimation with self-supervised learning to solve the scarce labeled data problem. Our self-supervised learning uses random cut-out and random normal noise as data augmentation to help the reconstruction pretext task for feature learning. The following downstream task predicts running and walking stride length and their type classification, and adopts a percentage error weighted RMSE to balance data difference of running and walking stride length. 
The simulation results show that the proposed method can achieve average error of 4.78\% on stride length and 99.83\% accuracy on running/walking classification, which outperforms the previous best result, 7.44\% average error on stride length. Further error reduction is possible by incorporating more data into this model.

% % \begin{thebibliography}{00}
\bibliographystyle{IEEEtran}
\bibliography{refs.bib}

% % \end{thebibliography}

\begin{IEEEbiography}[{\includegraphics[width=1in,height=1.25in,clip,keepaspectratio]{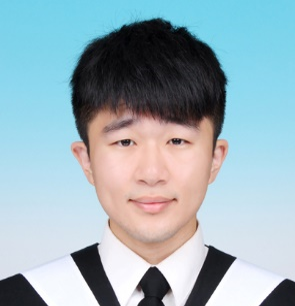}}]{Jien-De Sui} is curently pursuing the Ph.D. degree with the National Chiao Tung University (NCTU), Hsinchu, Taiwan, R.O.C., and focus on deep learning algorithm, IMU sensor system and VLSI design.
\end{IEEEbiography}

\begin{IEEEbiography}[{\includegraphics[width=1in,height=1.25in,clip,keepaspectratio]{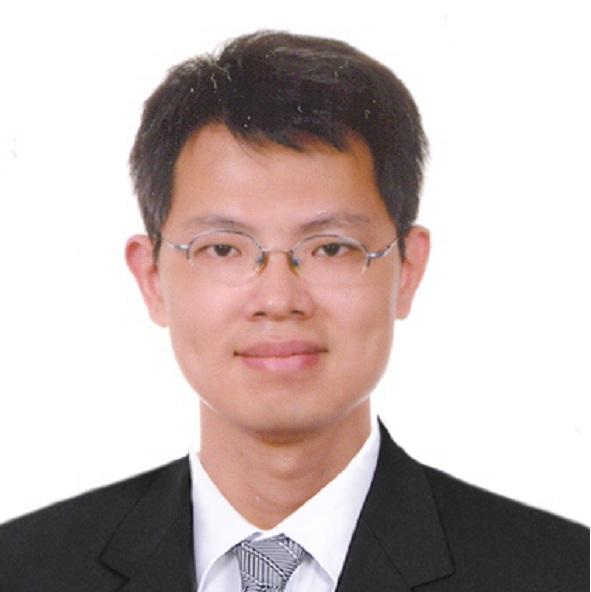}}]{Tian-Sheuan Chang} (S’93–M’06–SM’07)
	received the B.S., M.S., and Ph.D. degrees in electronic engineering from National Chiao-Tung University (NCTU), Hsinchu, Taiwan, in 1993, 1995, and 1999, respectively. 
	From 2000 to 2004, he was a Deputy Manager with Global Unichip Corporation, Hsinchu, Taiwan. In 2004, he joined the Department of Electronics Engineering, NCTU, where he is currently a Professor. In 2009, he was a visiting scholar in IMEC, Belgium. His current research interests include system-on-a-chip design, VLSI signal processing, and computer architecture.
	Dr. Chang has received the Excellent Young Electrical Engineer from Chinese Institute of Electrical Engineering in 2007, and the Outstanding Young Scholar from Taiwan IC Design Society in 2010. He has been actively involved in many international conferences as an organizing committee or technical program committee member.

\end{IEEEbiography}

\end{document}